# A Model-Based Approach to Rounding in Spectral Clustering

**Leonard K. M. Poon**    **April H. Liu**    **Tengfei Liu**    **Nevin L. Zhang**

Department of Computer Science and Engineering
The Hong Kong University of Science and Technology
Hong Kong, China
{lkmpoon, aprillh, liutf, lzhang}@cse.ust.hk

## Abstract

In spectral clustering, one defines a similarity matrix for a collection of data points, transforms the matrix to get the Laplacian matrix, finds the eigenvectors of the Laplacian matrix, and obtains a partition of the data using the leading eigenvectors. The last step is sometimes referred to as *rounding*, where one needs to decide how many leading eigenvectors to use, to determine the number of clusters, and to partition the data points. In this paper, we propose a novel method for rounding. The method differs from previous methods in three ways. First, we relax the assumption that the number of clusters equals the number of eigenvectors used. Second, when deciding the number of leading eigenvectors to use, we not only rely on information contained in the leading eigenvectors themselves, but also use subsequent eigenvectors. Third, our method is model-based and solves all the three subproblems of rounding using a class of graphical models called *latent tree models*. We evaluate our method on both synthetic and real-world data. The results show that our method works correctly in the ideal case where between-clusters similarity is 0, and degrades gracefully as one moves away from the ideal case.

## 1 Introduction

Clustering is a data analysis task where one assigns similar data points to the same cluster and dissimilar data points to different clusters. There are many different clustering algorithms, amongst which *spectral clustering* [14] is one approach that has gained prominence in recent years. The idea is to convert clustering into a graph cut problem. More specifically, one first builds a *similarity graph* over the data points using data similarity as edge weights and then partitions the graph by cutting some of the edges. Each connected component in the resulting graph is a cluster. The cut is done so as to simultaneously minimize the cost of cut and balance the sizes of the resulting clusters [12]. The graph cut problem is NP-hard and is hence relaxed. The solution is given by the leading eigenvectors of the so-called Laplacian matrix, which is a simple transformation of the original data similarity matrix. In a post-processing step called rounding [2], a partition of the data points is obtained from those real-valued eigenvectors.

In this paper we focus on rounding. In general, rounding is considered an open problem. There are three subproblems: (1) decide which leading eigenvectors to use; (2) determine the number of clusters; and (3) determine the members in each cluster. Previous rounding methods fall into two groups depending on whether they assume the number of clusters is given. When the number of clusters is known to be $k$, rounding is usually done based on the $k$ leading eigenvectors. The data points are projected onto the subspace spanned by those eigenvectors and then the K-means algorithm is run on that space to get $k$ clusters [14]. Bach and Jordan [2] approximate the subspace using a space spanned by $k$ piecewise constant vectors and then run K-means on the latter space. Zhang and Jordan [19] observe a link between rounding and the orthogonal Procrustes problem and iteratively use an analytical solution for the latter problem to build a method for rounding. Rebagliati and Verri [10] ask the user to provide a number $K$ that is larger than $k$ and obtain $k$ clusters based on the $K$ leading eigenvectors using a randomized algorithm.

When the number of clusters is not given, one needs to estimate it. A common method is to manually examine the difference between every two consecutive eigenvalues starting from the first two. If a big gap appears for the first time between the $k$-th and $(k+1)$- th eigenvalues, then one uses $k$ as an estimate of the number of clusters. Zelnik-Manor and Perona [17] propose an automatic method. For a particular integer $k$, it tries to rotate the $k$ leading eigenvectors so as to align them with the canonical coordinate system for the eigenspace spanned by those vectors. A cost function is defined in terms of how well the alignment can be achieved. The $k$ with the lowest cost is chosen as an estimate for the number of clusters. Xiang and Gong [16] and Zhao et al. [20] relax the assumption that clustering should be based on a continuous block of eigenvectors at the beginning of the eigenvector spectrum. They use heuristics to choose eigenvectors which do not necessarily form a continuous block, and use Gaussian mixture models to determine the clusters. Socher et al. [13] assume the number of leading eigenvectors to use is given. Based on those leading eigenvectors, they determine the number of clusters and the membership of each cluster using a non-parametric Bayesian clustering method.

In this paper, we propose a model-based approach to rounding. The novelty lies in three aspects. First, we relax the assumption that the number of clusters equals the number of eigenvectors used for rounding. While the assumption is reasonable in the *ideal case* where between-cluster similarity is 0, it is not appropriate in non-ideal cases. Our method allows the number of clusters to differ from the number of eigenvectors. Second, we choose a continuous block of leading eigenvectors for rounding just as [17]. The difference is that, when deciding the appropriateness of the $k$ leading eigenvectors, Zelnik-Manor and Perona [17] use only information contained in those eigenvectors, whereas we also use information in subsequent eigenvectors. So our method uses more information and hence the choice is expected to be more robust. Third, we solve all the three subproblems of rounding within one class of models, namely latent tree models [18]. In contrast, most previous methods either assume that the first two subproblems are solved and focus only on the third subproblem, or do not solve all the subproblems within one class of models [17, 20].

The remaining paper is organized as follows. In Section 2 we review the basics of spectral clustering and point out two key properties of the eigenvectors of the Laplacian matrix in the ideal case. In Section 3 we describe a straightforward method for rounding that takes advantage of the two properties. This method is fragile and breaks down as soon as we move away from the ideal case. In Sections 4 and 5 we propose a model-based method that exploits the same two properties. In Section 6 we evaluate the method on both synthetic and real-world data sets and also on real-world image segmentation tasks. The paper concludes in Section 7.

## 2 Background

To cluster a set of $n$ data points $X = \{x_1, \ldots, x_n\}$, one starts by defining a non-negative similarity measure $s_{ij}$ for each pair $x_i$ and $x_j$ of data points. In this paper we consider two measures. The *k-NN similarity measure* sets $s_{ij}$ to 1 if $x_i$ is one of the $k$ nearest neighbors of $x_j$, or vice verse, and 0 otherwise. The *Gaussian similarity measure* sets $s_{ij} = \exp(\frac{-\|x_i - x_j\|^2}{\sigma^2})$, where $\sigma$ controls the width of neighborhood of each data point. The $n \times n$ matrix $S = [s_{ij}]$ is called the *similarity matrix*. The *similarity graph* is an undirected graph over the data points where the edge is added between $x_i$ and $x_j$, with weight $s_{ij}$, if and only if $s_{ij} > 0$.

In spectral clustering, one transforms the similarity matrix to get the *Laplacian matrix*. There are several Laplacian matrices to choose from [14]. We use the *normalized Laplacian matrix* $L_{rw} = I - D^{-1}S$, where $I$ is the identity matrix, and $D = (d_{ij})$ is the diagonal *degree matrix* given by $d_{ii} = \sum_{j=1}^{n} s_{ij}$. Next one computes the eigenvectors of $L_{rw}$ and sorts them in ascending order of their corresponding eigenvalues. Denote those eigenvectors as $e_1, \ldots, e_n$. To obtain $k$ clusters, one forms an $n \times k$ matrix $U$ using the $k$ leading eigenvectors $e_1, \ldots, e_k$ as columns. Each row $y_i$ of $U$ is a point in $\mathbb{R}^k$ and corresponds to an original data point $x_i$. One partitions the points $\{y_1, \ldots, y_n\}$ into $k$ clusters using the K-means algorithm and obtains $k$ clusters of the original data points accordingly.

Suppose the data consist of $k_t$ true clusters $C_1, \ldots, C_{k_t}$. In the *ideal case*, the similarity graph has $k_t$ connected components, where each of them corresponds to a true cluster. Assume that the data points are ordered based on their cluster memberships. Then the Laplacian matrix $L_{rw}$ has a block diagonal form, with each block corresponding to one true cluster [14]. Moreover, $L_{rw}$ has exactly $k_t$ eigenvalues that equal 0, and the eigenspace of eigenvalue 0 is spanned by the indicator vectors $\vec{1}_{C_1}, \ldots, \vec{1}_{C_{k_t}}$. Create an $n \times k_t$ matrix $U_c$ using those indicator vectors as columns. It is known that, when $k = k_t$, $U = U_c R$ for some orthogonal matrix $R \in \mathbb{R}^{k \times k}$ [8]. For simplicity, assume the non-zero eigenvalues of $L_{rw}$ are distinct. Call those eigenvectors for eigenvalue 0 the *primary eigenvectors* and the others the *secondary eigenvectors*. The facts reviewed in this paragraph imply the following:

**Proposition 1.** *In the ideal case, eigenvectors of $L_{rw}$ have the following properties:*

1. *The primary eigenvectors are piecewise constant, with each value identifying one true cluster or the union of several true clusters.*
2. *The support (non-zero elements) of each secondary eigenvector is contained by one of the true clusters.*

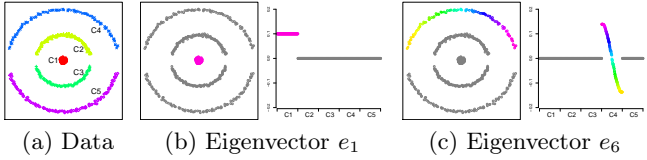

(a) Data  (b) Eigenvector $e_1$  (c) Eigenvector $e_6$

Figure 1: Example eigenvectors: There are five true clusters $C_1, \ldots, C_5$ in the data set shown in (a). The 10-NN similarity measure is used to produce the ideal case. Two eigenvectors are shown in (b) and (c). Each eigenvector is depicted in two diagrams. In the first diagram, the values of the eigenvector are indicated by different colors, with grey meaning 0. In the second diagram, the x-axis indexes the data points and the y-axis shows the values of the eigenvector.

The two properties are illustrated in Fig 1. The primary eigenvector $e_1$ has two different values, 0.1 and 0. The value 0.1 identifies the cluster $C_1$, whereas 0 corresponds to the union of the other clusters. The eigenvector $e_6$ is a secondary eigenvector. Its support is contained by the true cluster $C_4$.

## 3 A Naive Method for Rounding

In the original graph cut formulation of spectral clustering [12], an eigenvector has two different values. It partitions the data points into two clusters. In the relaxed problem, an eigenvector can have many different values (see Fig 1(c)). The first step of our method is to obtain, from each eigenvector $e_i$, two clusters using a confidence parameter $\delta$ that is between 0 and 1. Let $e_{ij}$ be the value of $e_i$ at the $j$-th data point. One of the clusters consists of the data points $x_j$ that satisfy $e_{ij} > 0$ and $e_{ij} > \delta \max_j e_{ij}$, while the other cluster consists of the data points $x_j$ that satisfy $e_{ij} < 0$ and $e_{ij} < \delta \min_j e_{ij}$. The indicator vectors of those two clusters are denoted as $e_i^+$ and $e_i^-$ respectively. We refer to the process of obtaining $e_i^+$ and $e_i^-$ from $e_i$ as *binarization*. Applying binarization to the eigenvectors $e_1$ and $e_6$ of Fig 1 results in the binary vectors shown in Fig 2. Note that $e_1^-$ is a degenerated binary vector in the sense that it is all 0. We still refer to it as a binary vector for convenience. The following proposition follows readily from Proposition 1.

**Proposition 2.** *Let $e^b$ be a vector obtained from an eigenvector $e$ of the Laplacian matrix $L_{rw}$ via binarization. In the ideal case:*

1. *If $e$ is a primary eigenvector, then the support of $e^b$ is either a true cluster or the union of several true clusters.*
2. *If $e$ is a secondary eigenvector, then the support of $e^b$ is contained by one of the true clusters.*

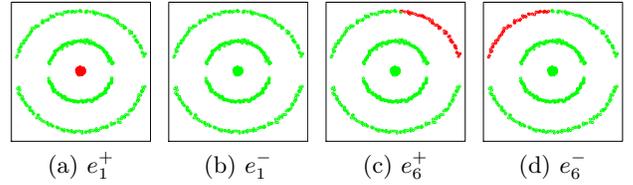

(a) $e_1^+$  (b) $e_1^-$  (c) $e_6^+$  (d) $e_6^-$

Figure 2: Binary vectors obtained from eigenvectors $e_1$ and $e_6$ through binarization with $\delta = 0.1$. Data points have values 1 (red) or 0 (green).

Each binary vector gives a partition of the data points, with one cluster comprising points with value 1 and another comprising points with value 0. Consider two partitions where one divides the data into two clusters $C_1$ and $C_2$ and the other into $P_1$ and $P_2$. Overlaying the two partitions results in a new partition consisting of clusters $C_1 \cap P_1$, $C_1 \cap P_2$, $C_2 \cap P_1$, and $C_2 \cap P_2$. Note that there are not necessarily exactly 4 clusters in the new partition as some of the 4 intersections might be empty. It is evident how to overlay multiple partitions.

Consider the case when the number $q$ of leading eigenvectors to use is given. Here is a straightforward method for rounding:

**NAIVE-ROUNDING1($q$)**

1. Compute the $q$ leading eigenvectors of $L_{rw}$.
2. Binarize the $q$ eigenvectors.
3. Obtain a partition of the data using each binary vector from the previous step.
4. Overlay all the partitions to get the final partition.

Note that the number of clusters obtained by NAIVE-ROUNDING1 is determined by $q$, but it is not necessarily the same as $q$. The following proposition is an easy corollary of Proposition 2.

**Proposition 3.** *In the ideal case and when $q$ is smaller than or equal to the number of true clusters, the clusters obtained by NAIVE-ROUNDING1 are either true clusters or unions of true clusters.*

Now consider how to determine the number $q$ of leading eigenvectors to use. We do so by making use of Proposition 2. The idea is to gradually increase $q$ in NAIVE-ROUNDING1 and test the partition $P_q$ obtained for each $q$ to see whether it satisfies the condition of Proposition 2 (2).

Suppose $K$ is a sufficiently large integer. Denote a subroutine that tests the partition $P_q$ by `cTest`($P_q, q, K$). To perform this test, we use the binary vectors obtained from the eigenvectors from the range $[q+1, K]$. If the support of every such binary vector is contained by some cluster in $P_q$, we say that $P_q$ satisfies the *containment condition* and `cTest` returns `true`. Oth-erwise, $P_q$ violates the condition and `cTest` returns `false`.

Let $k_t$ be the number of true clusters. When $q = k_t$, $\text{cTest}(P_q, q, K)$ passes because of Proposition 2. When $q = k_t + 1$, $P_q$ is likely to be finer than the true partition. Consequently, $\text{cTest}(P_q, q, K)$ may fail. The probability of this happening increases with the number of binary vectors used in the test. To make the probability high, we pick $K$ such that $K/2$ is safely larger than $k_t$ and let $q$ run from 2 to $\lfloor K/2 \rfloor$. When $q < k_t$, $\text{cTest}(P_q, q, K)$ usually passes.

The above discussions suggest that if $\text{cTest}(P_q, q, K)$, for the first time, passes for some $q = k$ and fails for $q = k + 1$, we can use $k$ as an estimate of $k_t$. Consequently, we can use the $k$ leading eigenvectors for clustering and return $P_k$ as the final clustering result. This leads to the following algorithm:

**NAIVE-ROUNDING2($K$)**

1. For $q = 2$ to $\lfloor K/2 \rfloor$,
   (a) $P \leftarrow $ NAIVE-ROUNDING1($q$).
   (b) $P' \leftarrow $ NAIVE-ROUNDING1($q + 1$).
   (c) If $\text{cTest}(P, q, K) = \text{true}$ and $\text{cTest}(P', q + 1, K) = \text{false}$, return $P$.
2. Return $P$.

Suppose an integer $K$ is given such that $K/2$ is safely larger than the number of true clusters. The algorithm NAIVE-ROUNDING2 automatically decides how many leading eigenvectors to use and automatically determines the number of clusters. Consider running it on the data set shown in Fig 1(a) with $K = 40$. It loops $q$ through 2 to 4 without terminating because both the two tests at Step 1(c) return `true`. When $q = 5$, $P$ is the true partition and $P'$ is shown in Fig 3(a). At Step 1(c), the first test for $P$ passes. However, the second test for $P'$ fails, because the support of the binary vector $e_{16}^+$ is not contained by any cluster in $P'$ (Fig 3). So, NAIVE-ROUNDING2 terminates at Step 1(c) and returns $P$ (the true partition) as the final result.

## 4 Latent Class Models for Rounding

NAIVE-ROUNDING2 is fragile. It can break down as soon as we move away from the ideal case. In this and the next section, we describe a model-based method that also exploits Proposition 2 and is more robust.

In the model-based method, the binary vectors are regarded as features and the problem is to cluster the data points based on those features. So what we face is a *problem of clustering discrete data*. As in the previous section, there is a question of how many leading eigenvectors to use. In this section, we assume the number $q$ of leading eigenvectors to use is given. In the next section, we will discuss how to determine $q$.

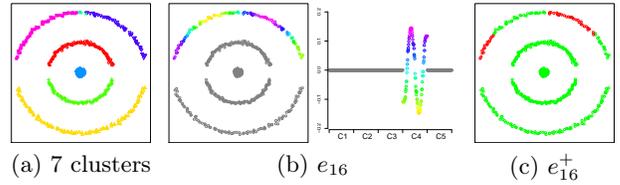

(a) 7 clusters    (b) $e_{16}$    (c) $e_{16}^+$

Figure 3: Illustration of NAIVE-ROUNDING2: (a) shows the 7 clusters given by the first 6 pairs of binary vectors. (b) and (c) show eigenvector $e_{16}$ and one of the binary vectors obtained from $e_{16}$. The support of $e_{16}^+$ (the red region) is not contained by any of the clusters in (a).

The problem we address in this section is how to cluster the data points based on the first $q$ pairs of binary vectors $e_1^+, e_1^-, \ldots, e_q^+, e_q^-$. We solve the problem using latent class models (LCMs) [3]. LCMs are commonly used to cluster discrete data, just as Gaussian mixture models are used to cluster continuous data. Technically they are the same as the Naive Bayes model except that the class variable is not observed.

The LCM for our problem is shown in Fig 4(a). So far we have been using the notations $e_s^+$ and $e_s^-$ to denote vectors of $n$ elements or functions over the data points. In this and the next sections, we overload the notations to denote random variables that take different values at different data points. The latent variable $Y$ represents the partition to find and each state of $Y$ represents a cluster. So the number of states of $Y$, often called the *cardinality* of $Y$, is the number of clusters. To learn an LCM from data means to determine: (1) the cardinality of $Y$, i.e., the number of clusters; and (2) the probability distributions $P(Y)$, $P(e_s^+|Y)$, and $P(e_s^-|Y)$, i.e., the characterization of the clusters.

After an LCM is learned, one can compute the posterior distribution of $Y$ for each data point. This gives a *soft partition* of the data. To get a *hard partition*, one can assign each data point to the state of $Y$ that has the maximum posterior probability. This is called *hard assignment*.

There are two cases with the LCM learning problem, depending on whether the number of clusters is known. When that number is known, we only need to determine the probability distributions $P(Y)$, $P(e_s^+|Y)$, and $P(e_s^-|Y)$. This is done using the EM algorithm [5].

**Proposition 4.** *It is possible to set the probability parameter values of the LCM in Fig 4(a) in such a way that it gives the same partition as* NAIVE-ROUDING1. *Moreover, those parameter values maximize the likelihood of the model.*

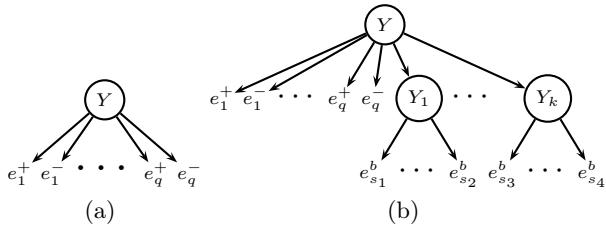

Figure 4: (a) Latent class model and (b) latent tree model for rounding: The binary vectors $e_1^+$, $e_1^-$, ..., $e_q^+$, $e_q^-$ are regarded as random variables. The discrete latent variable $Y$ represents the partition to find.

The proof is omitted due to space limit. It is well known that the EM algorithm aims at finding the maximum likelihood estimate (MLE) of the parameters. So the LCM method for rounding actually tries to find the same partition as NAIVE-ROUNDING1.

Now consider the case when the number of clusters is not known. We determine it using the BIC score [11]. Given a model $m$, the score is defined as $BIC(m) = \log P(D|m, \theta) - \frac{d}{2} \log n$, where $D$ is the data, $\theta$ is the MLE of the parameters, and $d$ is the number of free parameters in the model. We start by setting the number $k$ of clusters to 2 and increase it gradually. We stop the process as soon as the BIC score of the model starts to decrease, and use the $k$ with the maximum BIC score as the estimate of the number of clusters. Note that the number of clusters determined using the BIC score might not be the same as the number of clusters found by NAIVE-ROUNDING1.

## 5 Latent Tree Models for Rounding

In this section we present a method for determining the number $q$ of leading eigenvectors to use. Consider an integer $q$ between 2 and $K/2$. We first build an LCM using the first $q$ pairs of binary vectors and obtain a hard partition of the data using the LCM. Suppose $k$ clusters $C_1$, ..., $C_k$ are obtained. Each cluster $C_r$ corresponds to a state $r$ of the latent variable $Y$.

We extend the LCM model to obtain the model shown in Fig 4(b). We do this in three steps. First, we introduce $k$ new latent variables $Y_1$, ..., $Y_k$ into the model and connect them to $Y$. Each $Y_r$ is a binary variable and its conditional distribution is set as follows:

$$P(Y_r = 1 | Y = r') = \begin{cases} 1 & \text{if } r' = r, \\ 0 & \text{otherwise.} \end{cases} \quad (1)$$

So the state $Y_r = 1$ means the cluster $C_r$ and $Y_r = 0$ means the union of all the other clusters.

Next, we add binary vectors from the range $[q+1, K]$ to the model by connecting them to the new latent variables. For convenience we call those vectors *secondary binary vectors*. This is not to be confused with the secondary eigenvectors mentioned in Proposition 1. For each secondary binary vector $e_s^b$, let $D_s$ be its support. When determining to which $Y_r$ to connect $e_s^b$, we consider how well the cluster $C_r$ covers $D_s$. We connect $e_s^b$ to the $Y_r$ such that $C_r$ covers $D_s$ the best, in the sense that the quantity $|D_s \cap C_r|$ is maximized, where $|.|$ stands for the number of data points in a set. We break ties arbitrarily.

Finally, we set the conditional distribution $P(e_s^b | Y_r)$ as follows:

$$P(e_s^b = 1 | Y_r = 1) = \frac{|D_s \cap C_r|}{|C_r|} \quad (2)$$

$$P(e_s^b = 1 | Y_r = 0) = \frac{|D_s - C_r|}{n - |C_r|} \quad (3)$$

What we get after the extension is a tree structured probabilistic graphical model, in which the variables at the leaf nodes are observed and the variables at the internal nodes are latent. Such models are known as *latent tree models (LTMs)* [4, 15], sometimes also called hierarchical latent class models [18]. The LCM part of the model is called its *primary part*, while the newly added part is called the *secondary part*. The parameter values for the primary part is determined during LCM learning, while those for the secondary part are set manually by Equations (1–3).

To determine the number $q$ of leading eigenvectors to use, we examine all integers in the range $[2, K/2]$. For each such integer $q$, we build an LTM as described above and compute its BIC score. We pick the $q$ with the maximum BIC score as the answer. After $q$ is determined, we use the primary part of the LTM to partition the data. In other words, the secondary part is used only to determine $q$. We call this method *LTM-Rounding*.

Here are the intuitions behind the LTM method. If the support $D_s$ of $e_s^b$ is contained in cluster $C_r$, it fits the situation to connect $e_s^b$ to $Y_r$. The model construction no longer fits the data well if $D_s$ is not contained in any of the clusters. The worst case is when two different clusters $C_r$ and $C_{r'}$ cover $D_s$ equally well and better than other clusters. In this case, $e_s^b$ can be either connected to $Y_r$ or to $Y_{r'}$. Different choices here lead to different models. As such, neither choice is 'ideal'. Even when there is only one cluster $C_r$ that covers $D_s$ the best, connecting $e_s^b$ to $Y_r$ is still intuitively not 'perfect' as long as $C_r$ does not cover $D_s$ completely.

So when the support of every secondary binary vector is contained by one of the clusters $C_1$, ..., $C_k$, the

LTM we build would fit the data well. However, when the supports of some secondary binary vectors are not completely covered by any of the clusters, the LTM would not fit the data well. In the ideal case, the fit would be good if $q$ is the number $k_t$ of eigenvectors for eigenvalue 0 according to Proposition 2. The fit would not be as good otherwise. This is why the likelihood of the LTM contains information that can be used to choose $q$.

We now summarize the LTM method for rounding:

**LTM-ROUNDING($K,\delta$)**

1. Compute the $K$ leading eigenvectors of $L_{rw}$.
2. Binarize the eigenvectors with confidence parameter $\delta$ as in Section 3.
3. $S^* \leftarrow -\infty$.
4. For $q = 2$ to $\lfloor K/2 \rfloor$,
   (a) $m_{lcm} \leftarrow$ the LCM learnt using the first $q$ pairs of binary vectors as shown in Section 4.
   (b) $P \leftarrow$ hard partition obtained using $m_{lcm}$.
   (c) $m_{ltm} \leftarrow$ the LTM extended from $m_{lcm}$ as explained in Section 5.
   (d) $S \leftarrow$ the BIC score of $m_{ltm}$.
   (e) If $S > S^*$, then $P^* \leftarrow P$ and $S^* \leftarrow S$.
5. Return $P^*$.

An implementation of LTM-ROUNDING can be obtained from http://www.cse.ust.hk/~lzhang/ltm/index.htm.

The choice of $K$ should be such that $K/2$ is safely larger than the number of true clusters. We do not allow $q$ to be larger than $K/2$, so that there is sufficient information in the secondary part of the LTM to determine the appropriateness of using the $q$ leading eigenvectors for rounding. The parameter $\delta$ should be picked from the range $(0,1)$. We recommend to set $\delta$ at 0.1. The sensitivity of LTM-ROUNDING with respect to those parameters will be investigated in Section 6.3.

## 6 Empirical Evaluations

Our empirical investigations are designed to: (1) show that LTM-ROUNDING works perfectly in the ideal case and its performance degrades gracefully as we move away from the ideal case, and (2) compare LTM-ROUNDING with alternative methods. Synthetic data are used for the first purpose, while both synthetic and real-world data are used for the second purpose.

LTM-ROUNDING has two parameters $\delta$ and $K$. We set $\delta = 0.1$ and $K = 40$ in all our experiments except in sensitivity analysis (Section 6.3). For each set of synthetic data, 10 repetitions were run.

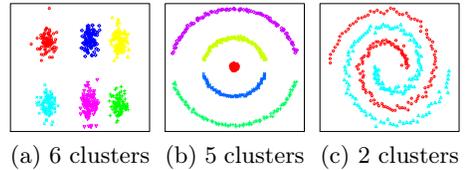

(a) 6 clusters  (b) 5 clusters  (c) 2 clusters

Figure 5: Synthetic data set for the ideal case. Each color and shape of the data points identifies a cluster. Clusters were recovered by LTM-ROUNDING correctly.

### 6.1 Performance in the Ideal Case

Three data sets were used for the ideal case (Fig 5). They vary in the number and the shape of clusters. Intuitively the first data set is the easiest, while the third one is the hardest. To produce the ideal case, we used the 10-NN similarity measure for the first two data sets. For the third one, the same measure gave a similarity graph with one single connected component. So we used the 3-NN similarity measure instead.

LTM-ROUNDING produced the same results on all 10 runs. The results are shown in Fig 5 using colors and shapes of data points. Each color identifies a cluster. LTM-ROUNDING correctly determined the numbers of clusters and recovered the true clusters perfectly.

### 6.2 Graceful Degrading of Performance

To demonstrate that the performance of LTM-ROUNDING degrades gracefully as we move away from the ideal case, we generated 8 new data sets by adding different levels of noise to the second data set in Fig 5. The Gaussian similarity measure was adopted with $\sigma = 0.2$. So the similarity graphs for all the data sets are complete.

We evaluated the quality of an obtained partition by comparing it with the true partition using Rand index (RI) [9] and variation of information (VI) [7]. Note that higher RI values indicate better performance, while the opposite is true for VI. The performance statistics of LTM-ROUNDING are shown in Table 1. We see that RI is 1 for the first three data sets. This means that the true clusters have been perfectly recovered. The index starts to drop from the 4th data set onwards in general. It falls gracefully with the increase in the level of noise in data. Similar trend can also be observed on VI.

The partitions produced by LTM-ROUNDING at the best run (in terms of BIC score) are shown in Fig 7(a) using colors and shapes of the data points. We see that on the first four data sets, LTM-ROUNDING correctly determined the number of clusters and the members of each cluster. On the next three data sets, it also

Table 1: Performances of various methods on the 8 synthetic data sets in terms of Rand index (RI) and variation of information (VI). Higher values of RI or lower values of VI indicate better performance. K-MEANS and GMM require extra information for rounding and should not be compared with the other two methods directly.

|    | Data    | 1        | 2        | 3        | 4        | 5        | 6        | 7        | 8        |
|----|---------|----------|----------|----------|----------|----------|----------|----------|----------|
| RI | LTM     | **1.0±.00** | **1.0±.00** | **1.0±.00** | **.99±.01** | **.97±.02** | **.98±.01** | **.94±.01** | .88±.01 |
|    | ROT     | .92±.00  | .92±.00  | 1.0±.00  | .98±.00  | .52±.00  | .52±.00  | .88±.00  | **.90±.00** |
|    | K-MEANS | 1.0±.00  | 1.0±.00  | 1.0±.00  | 1.0±.00  | .85±.00  | .72±.00  | .71±.00  | .75±.00  |
|    | GMM     | 1.0±.00  | 1.0±.00  | 1.0±.00  | 1.0±.00  | .94±.00  | .88±.00  | .91±.00  | .88±.00  |
| VI | LTM     | **.00±.00** | **.00±.00** | **.00±.00** | **.06±.09** | **.29±.19** | **.28±.14** | **.79±.12** | 1.64±.10 |
|    | ROT     | .40±.00  | .40±.00  | .00±.00  | .20±.00  | 1.60±.00 | 1.60±.00 | 1.85±.00 | **1.42±.00** |
|    | K-MEANS | .00±.00  | .00±.00  | .00±.00  | .00±.00  | 1.04±.00 | 1.41±.00 | 1.52±.00 | 1.97±.00 |
|    | GMM     | .00±.00  | .00±.00  | .00±.00  | .00±.00  | .85±.00  | .91±.00  | 1.25±.00 | 1.74±.00 |

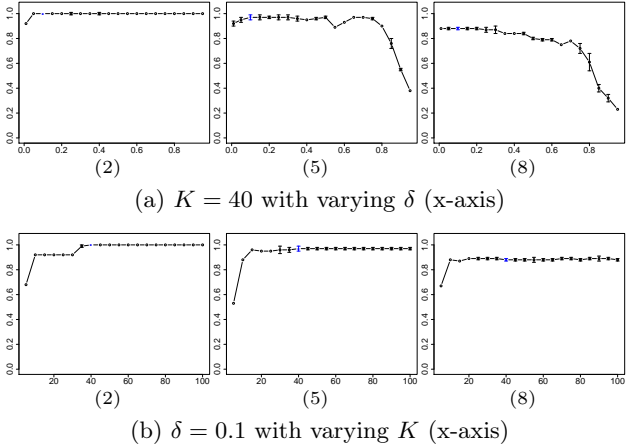

Figure 6: Sensitivity analysis on parameters $\delta$ and $K$ in LTM-ROUNDING. The y-axis denotes Rand index. We recommend $\delta = 0.1$ and $K = 40$ in general.

(a) $K = 40$ with varying $\delta$ (x-axis)

(b) $\delta = 0.1$ with varying $K$ (x-axis)

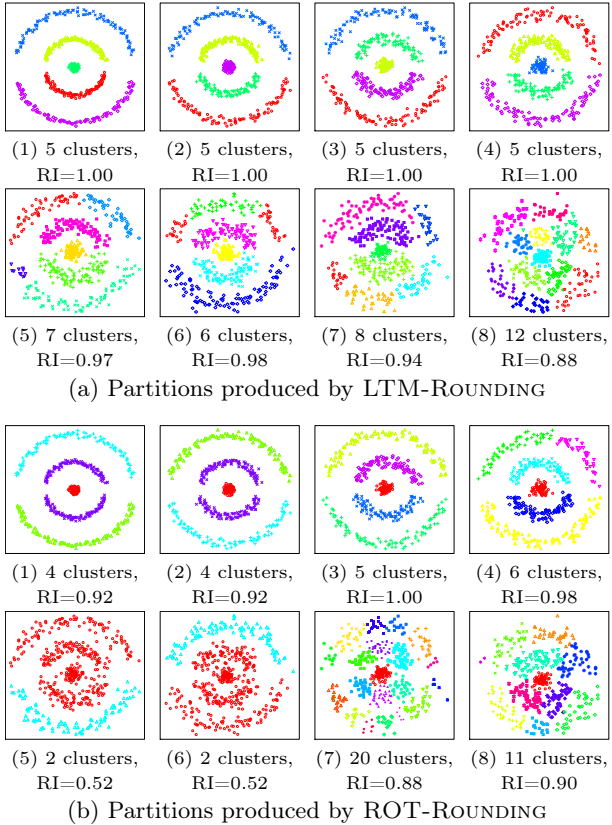

Figure 7: The 8 synthetic data sets.

(a) Partitions produced by LTM-ROUNDING

(b) Partitions produced by ROT-ROUNDING

correctly recovered the true clusters except that the top and the bottom crescent clusters might have been broken into two or three parts. Given the gaps between the different parts, the results are probably the best one can hope for. The result on the last data set is less ideal but still reasonable.

Putting together, the above discussions indicate that the performance of LTM-ROUNDING degrades gracefully as we move away from the ideal case.

### 6.3 Sensitivity Analysis

We now study the sensitivity of the parameters $\delta$ and $K$ in LTM-ROUNDING. Experiments were conducted on the 2nd, 5th, and 8th data sets in Fig 7. Those data sets contain different levels of noise and hence are at different distances from the ideal case.

To determine the sensitivity of LTM-ROUNDING with respect to $\delta$, we fix $K = 40$ and let $\delta$ vary between 0.01 and 0.95. The RI statistics are shown in Fig 6(a). We see that on data set (2) the performance of LTM-ROUNDING is insensitive to $\delta$ except when $\delta = 0.01$.

On data set (5), the performance of LTM-ROUNDING is more or less robust when $0.1 \leq \delta \leq 0.3$. Its performance gets particularly worse when $\delta \geq 0.8$. Similar behavior can be observed on data set (8). Those results suggest that the performance of LTM-ROUNDING is robust with respect to $\delta$ in situations close to the ideal case and it becomes sensitive in situations that are far away from the ideal case. In general, we recommend $\delta = 0.1$.

To determine the sensitivity of LTM-ROUNDING with respect to $K$, we fix $\delta = 0.1$ and let $K$ vary between 5 and 100. The RI statistics are shown in Fig 6(b). It is clear that the performance of LTM-ROUNDING is robust with respect to $K$ as long as it is not too small.

### 6.4 Running Time

LTM-ROUNDING deals with tree-structured models. It is deterministic everywhere except at Step 4(a), where the EM algorithm is called to estimate the parameters of LCM. EM is an iterative algorithm and is computationally expensive in general. However, it is efficient on LCMs. So the running time of LTM-ROUNDING is relatively short. To process one of the data sets in Fig 7, it took around 10 minutes on a laptop computer.

### 6.5 Alternative Methods for Rounding

To compare with LTM-ROUNDING, we included the method by Zelnik-Manor and Perona [17] in our experiments. The latter method determines the appropriateness of using the $q$ leading eigenvectors by checking how well they can be aligned with the canonical coordinates through rotation. So we name it ROT-ROUNDING. Both LTM-ROUNDING and ROT-ROUNDING can determine the number $q$ of eigenvectors and the number $k$ of clusters automatically. Therefore, they are directly comparable. The method by Xiang and Gong [16] is also related to our work. However, its implementation is not available to us and hence it is excluded from comparison.

K-MEANS and GMM can also be used for rounding. However, both methods cannot determine $q$, and K-MEANS cannot even determine $k$. We therefore gave the number $k_t$ of true clusters for K-MEANS and used $q = k_t$ for both methods. Since the two methods required additional information, they are included in our experiments only for reference. They should not be compared with LTM-ROUNDING directly.

ROT-ROUNDING and GMM require the maximum allowable number of clusters. We set that number to 20.

### 6.6 Comparison on Synthetic Data

Table 1 shows the performance statistics of the three other methods on the 8 synthetic data sets. LTM-ROUNDING performed better than ROT-ROUNDING on all but two data sets. The differences were substantial on 5 of those data sets.

The partitions produced by ROT-ROUNDING at one run are shown in Fig 7(b).[1] We see that on the first two data sets, ROT-ROUNDING underestimated the number of clusters and merged the two smaller crescent clusters. This is serious under-performance given the easiness of those data sets. On the 3rd data set, it recovered the true clusters correctly. On the 4th data set, it broke the top crescent cluster into two. On the 5th data set, it recovered the bottom cluster correctly but merged all the other clusters incorrectly. This leads to a much lower RI. The story on the 6th data set is similar. On the 7th data set, it produced many more clusters than the number of true clusters. Therefore, its RI is also lower. On the last data set, the clustering obtained by ROT-ROUNDING is not visually better than the one produced by LTM-ROUNDING, even though RI suggests otherwise. Given all the evidence presented, we conclude that the performance of LTM-ROUNDING is significantly better than that of ROT-ROUNDING.

---

[1]Same results were obtained for all 10 repetitions.

Table 2: Comparison on MNIST digits data.

| Method  | #clusters | RI      | VI        |
|---------|-----------|---------|-----------|
| LTM     | **9.3**±.82 | .91±.00 | **2.17**±.07 |
| ROT     | 19.0±.00  | **.92**±.00 | 2.40±.00  |
| K-MEANS | (10)      | .90±.00 | 1.83±.00  |
| GMM     | 16.0±.00  | .91±.00 | 2.14±.00  |
| SD-CRP  | 9.3±.96   | .89±.00 | 2.72±.08  |

Like LTM-ROUNDING, both K-MEANS and GMM recovered clusterings correctly on the first three data sets (Table 1). They performed slightly better than LTM-ROUNDING on the 4th data set. However, their performances were considerably worse on the next three data sets. They were also worse on the last one in terms of VI. It happened even though K-MEANS and GMM were given additional information for rounding. This shows the superior performance of LTM-ROUNDING.

### 6.7 MNIST Digits Data

In the next two experiments, we used real-world data to compare the rounding methods. The MNIST digits data were used in this subsection. The data consist of 1000 samples of handwritten digits from 0 to 9. They were preprocessed by the deep belief network as described in [6] using their accompanying code.[2] Table 2 shows the results averaged over 10 runs.

We see that the number of clusters estimated by LTM-ROUNDING is close to the ground truth (10), but that by ROT-ROUNDING is considerably larger than 10. In terms of quality of clusterings, the results are inconclusive. RI suggests that ROT-ROUNDING performed slightly better, but VI suggests that LTM-ROUNDING performed significantly better.

Compared with LTM-ROUNDING, K-MEANS obtained

---

[2]We thank Richard Socher for sharing the preprocessed data with us. The original data can be found at http://yann.lecun.com/exdb/mnist/.

a better clustering (in terms of VI), whereas GMM obtained one with similar quality. However, K-MEANS was given the number of true clusters and GMM the number of eigenvectors. GMM also overestimated the number of clusters even with the extra information.

A non-parametric Bayesian clustering method, called SD-CRP, has recently been proposed for rounding [13]. Although we obtained the same data from their authors, we could not get a working implementation for their method. Therefore, we simply copied their reported performance to Table 2. Note that SD-CPR can determine the number clusters automatically. However, it requires the number of eigenvectors as input, and the number was set to 10 in this experiment. Hence, SD-CPR requires more information than LTM-ROUNDING. Table 2 shows that it estimated a similar number of clusters as LTM-ROUNDING. However, its clustering was significantly worse in terms of RI or VI. This shows that LTM-ROUNDING performed better than SD-CPR even with less given information.

### 6.8 Image Segmentation

The last part of our empirical evaluation was conducted on real-world image segmentation tasks. Five images from the Berkeley Segmentation Data Set (BSDS500) were used. They are shown in the first column of Fig 8. The similarity matrices were built using the method proposed by Arbeláez et al. [1].

The segmentation results obtained by ROT-ROUNDING and LTM-ROUNDING are shown in the second and third columns of Fig 8 respectively. On the first two images, ROT-ROUNDING did not identify any meaningful segments. In contrast, LTM-ROUNDING identified the polar bear and detected the boundaries of the river on the first image. It identified the bottle, the glass and a lobster on the second image.

An obvious undesirable aspect of the results is that some uniform regions are broken up. Examples include the river bank and the river itself in the first image, and the background and the table in the second image. This is a known problem of spectral clustering when applied to image segmentation and can be dealt with using image analysis techniques [1]. We do not deal with the problem in this paper.

On the third image the performance of LTM-ROUNDING was better because the lizard was identified and the segmentation lines follow leaf edges more closely. On the fourth image LTM-ROUNDING did a better job at detecting the edges around the lady's hands and skirt and on the left end of the silk scarf. However, ROT-ROUNDING did a better job at the last image because it produced a cleaner segmentation.

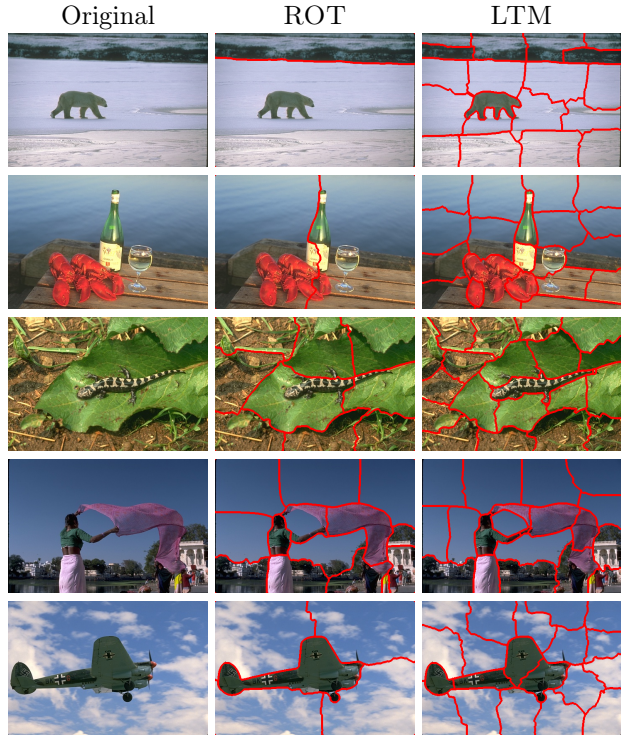

Figure 8: Image segmentation results.

Overall, the performance of LTM-ROUNDING was better than that of ROT-ROUNDING. The objective of this section has been to compare LTM-ROUNDING and ROT-ROUNDING on the same collection of eigenvectors. The conclusion is meaningful even if the final segmentation results are not as good as the best that can be achieved by image analysis techniques.

## 7 Conclusion

Rounding is an important step of spectral clustering that has not received sufficient attention. Not many papers have been published on the topic, especially on the issues of determining the number of leading eigenvectors to use. In this paper, we have proposed a novel method for the task. The method is based on latent tree models. It can automatically select an appropriate number of eigenvectors to use, determine the number of clusters, and finally assign data points to clusters. We have shown that the method works correctly in the ideal case and its performance degrades gracefully as we move away from the ideal case.

**Acknowledgements**

Research on this paper was supported by China National Basic Research 973 Program project No. 2011CB505101 and Guangzhou HKUST Fok Ying Tung Research Institute.